\documentclass[10pt,twocolumn,letterpaper]{article}

\usepackage{multirow}
\usepackage{bbding}
\usepackage{graphicx}
\usepackage{subfig}
\usepackage{float}
\usepackage{caption}
\usepackage[accsupp]{axessibility}  % Improves PDF readability for those with disabilities.

\usepackage[pagenumbers]{cvpr} % To force page numbers, e.g. for an arXiv version

\definecolor{cvprblue}{rgb}{0.21,0.49,0.74}
\usepackage[pagebackref,breaklinks,colorlinks,allcolors=cvprblue]{hyperref}

\def\paperID{5367} % *** Enter the Paper ID here
\def\confName{CVPR}
\def\confYear{2025}

\title{Taste More, Taste Better: Diverse Data and Strong Model \\ Boost Semi-Supervised Crowd Counting}

\author{%
Maochen Yang$^{1}$ \quad 
Zekun Li$^{1}$ \quad 
Jian Zhang$^{1}$ \quad
Lei Qi$^{3}$ \quad
Yinghuan Shi$^{1, 2, *}$ \\
$^{1}$Nanjing University, China \quad
$^{2}$Suzhou Laboratory, China \quad
$^{3}$Southeast University, China \\
{\tt\small \{yangmaochen, lizekun, zhangjian7369\}@smail.nju.edu.cn} \quad
{\tt\small qilei@seu.edu.cn, syh@nju.edu.cn}
}

\def\tablegap{-4 mm}

\begin{document}
\maketitle

\begin{abstract}
Semi-supervised crowd counting is crucial for addressing the high annotation costs of densely populated scenes. Although several methods based on pseudo-labeling have been proposed, it remains challenging to effectively and accurately utilize unlabeled data. In this paper, we propose a novel framework called Taste More Taste Better (TMTB), which emphasizes both data and model aspects. Firstly, we explore a data augmentation technique well-suited for the crowd counting task. By inpainting the background regions, this technique can effectively enhance data diversity while preserving the fidelity of the entire scenes. Secondly, we introduce the Visual State Space Model as backbone to capture the global context information from crowd scenes, which is crucial for extremely crowded, low-light, and adverse weather scenarios. In addition to the traditional regression head for exact prediction, we employ an Anti-Noise classification head to provide less exact but more accurate supervision, since the regression head is sensitive to noise in manual annotations. We conduct extensive experiments on four benchmark datasets and show that our method outperforms state-of-the-art methods by a large margin. Code is publicly available on \href{https://github.com/syhien/taste_more_taste_better}{https://github.com/syhien/taste\_more\_taste\_better}. 
\end{abstract}
    
\let\thefootnote\relax\footnote{* Corresponding author: Yinghuan Shi. This work was supported by National Science and Technology Major Project (2023ZD0120700), NSFC Project (62222604, 62206052, 624B2063), China Postdoctoral Science Foundation (2024M750424), Fundamental Research Funds for the Central Universities (020214380120, 020214380128), State Key Laboratory Fund (ZZKT2024A14), Postdoctoral Fellowship Program of CPSF (GZC20240252), Jiangsu Funding Program for Excellent Postdoctoral Talent (2024ZB242), and Jiangsu Science and Technology Major Project (BG2024031).}
\begin{figure}
    \centering
    \subfloat{\includegraphics[width=0.85\linewidth]{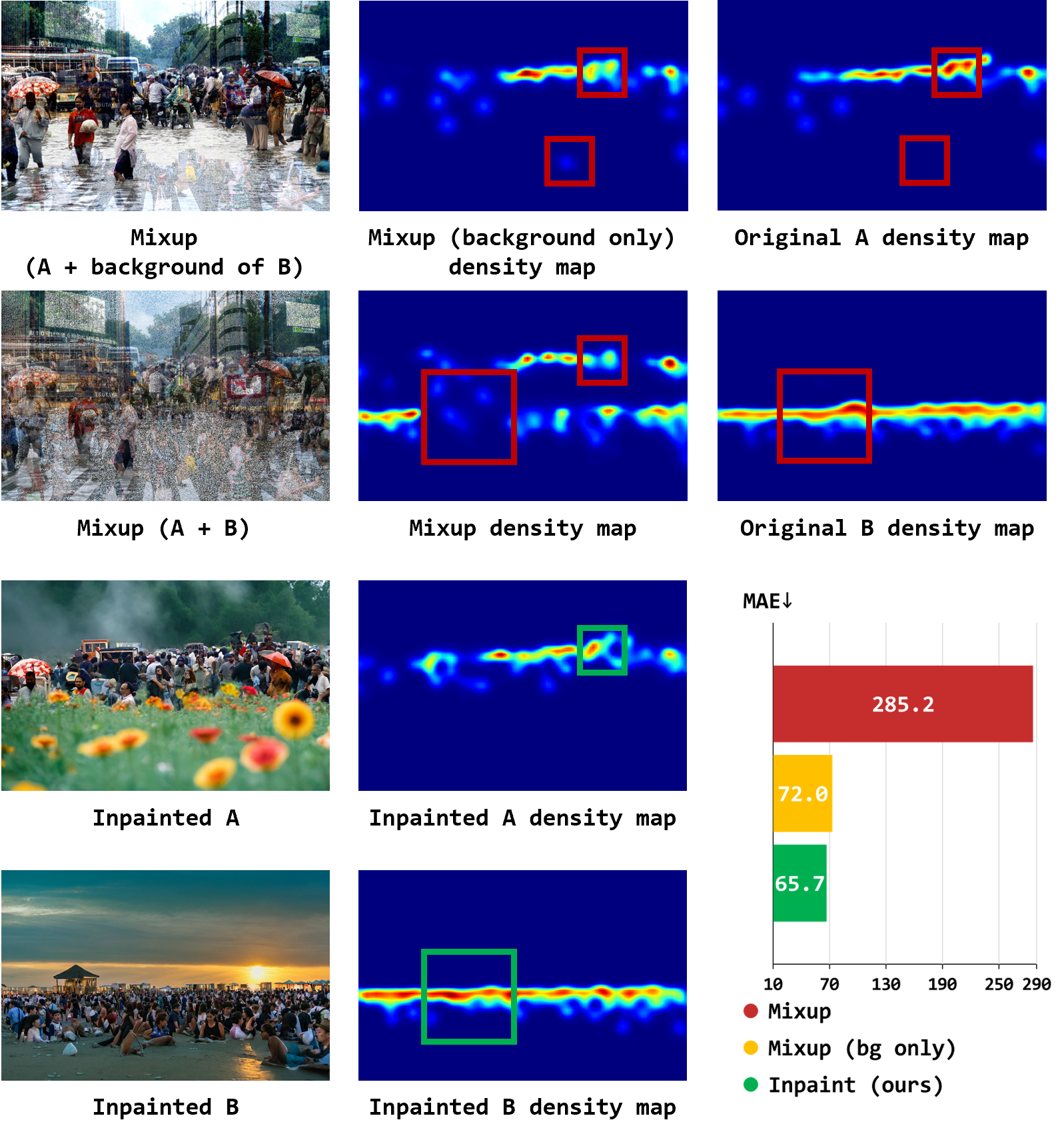}}
    \vspace{-0.2 cm}
    \caption{Comparisons of augmentations. As shown in the \textcolor{red}{red} bounding boxes, both types of Mixup generate disappointing density maps. Mixup is supposed to retain crowd information from two images, but actually it destroys the spatial structure. As shown in \textcolor{green}{green} bounding boxes, our inpainting augmentation is better suited for the crowd counting task, showing an impressive reduction in Mean Absolute Error (MAE).}
    \label{fig:intro_aug}
    \vspace{-0.4 cm}
\end{figure}

\section{Introduction}

\label{sec:intro}

Crowd counting is a fundamental task in computer vision that involves estimating the number of people in densely populated scenes. This task is of great importance in various public safety scenarios, including transport management \cite{handte2014crowd,soy2023edge,a_crowdconting_in_public_transportation} and disaster prevention \cite{sun2009real}. The substantial demand has driven the advancement of counting methods \cite{zhang2015cross,boominathan2016crowdnet,ranjan2018iterative,liu2019context,shi2019revisiting}. 
However, a significant cost of manual annotation is unavoidable for these fully-supervised deep methods to achieve satisfactory performance. For example, the UCF-QNRF dataset with $1,535$ images and $1,251,642$ annotations amounts to approximately $2,000$ human-hours \cite{ucfqnrf}. 
Compared to the extremely expensive labeled data with annotations, acquiring unlabeled crowd images is much more affordable. Consequently, there is a growing interest in leveraging unlabeled data through semi-supervised crowd counting (SSCC) methods. The recent mainstream works on SSCC \cite{lin2023optimal,Calibrating,zhang2024boosting,mrc-crowd2024} mostly employ the Mean Teacher \cite{mean_teacher} framework, where the teacher model generates pseudo-labels for unlabeled data to guide the student model. The key to achieving superior performance with SSCC methods on a given set of images, where only a small portion is annotated, lies in how we handle the unlabeled samples and how we design an appropriate model to fully exploit the information within them.

It has been observed that introducing similar but different samples to the training data can enlarge the support of the training distribution and thus enhance the robustness and generalizability of the model \cite{Simard2012,vrm}. In the literature, numerous data augmentation techniques have demonstrated their effectiveness in generating additional virtual samples to support the training process in both fully-supervised and semi-supervised settings \cite{zhang2018mixupempiricalriskminimization,yun2019cutmixregularizationstrategytrain,bai2023bidirectionalcopypastesemisupervisedmedical}. However, we find that these off-the-shelf techniques cannot suit the crowd counting task, as they may destroy the spatial structure of crowd scenes, leading to distorted density maps, as shown in \Cref{fig:intro_aug}. Therefore, a well-suited data augmentation technique specifically designed for the crowd counting task is crucial for the model to \textbf{\textit{taste more}} from the unlabeled data.

With the help of appropriate data augmentation, the diversity of unlabeled data during training can be significantly improved. However, this simultaneously brings new challenges for the teacher model to produce correct supervision signals. From the perspective of the model itself, we find that the Convolutional Neural Network (CNN) architecture commonly used in previous studies can be suboptimal for the crowd counting task, as it tends to overfit to local details and fails to capture global context information, which is crucial for extremely crowded, low-light, and adverse weather scenarios. From another perspective, it is also important to generate more accurate pseudo-labels while filtering out low-quality parts. Considering these two points, we explore a modern backbone network to replace CNNs and enhance it with a holistic strategy for producing and leveraging pseudo-labels, which helps to \textbf{\textit{taste}} data \textbf{\textit{better}}.

In this paper, we propose a novel semi-supervised crowd counting method, named \textbf{\textit{Taste More Taste Better}} (TMTB). 
We design a novel data augmentation technique, Inpainting Augmentation, which enhances the diversity of the dataset while keeping the integrity of the foreground regions. 
Observing that some regions in inpainted images are noisy and harmful, to better utilize the inpainted images, we employ a Visual State Space Model (VSSM) with an Anti-Noise classification branch for assistance, which provides inexact but accurate supervision. 
The proposed model generates more accurate density maps than state-of-the-art methods. 
Our method TMTB achieves significant improvements on several benchmark datasets, including ShanghaiTech Part A \& B~\cite{shanghaiAB}, UCF-QNRF~\cite{ucfqnrf}, and JHU-Crowd++~\cite{sindagi2020jhu-crowd++}. We reduce the Mean Absolute Error of JHU-Crowd with 5\% data labeled to 67.0, marking the first time it has been lower than 70.0. 
The method also presents strong generalization ability in cross-dataset evaluations. 

The main contributions of this paper are as follows:
\begin{itemize}
    \item We observed that the lack of suitable data augmentation methods and the weakness of capturing global context information are bottlenecks for the crowd counting task. The off-the-shelf data methods may destroy the spatial structure of crowd scenes, which is crucial for the regression of density maps. CNNs tend to overfit local details and struggle to capture global context information. 
    \item We propose a novel semi-supervised crowd counting method. A data augmentation technique is designed to enhance diversity while keeping the integrity of the foreground regions. A VSSM with an Anti-Noise classification branch for assistance is employed, which provides inexact but accurate supervision, to better utilize the data.
    \item The proposed method remarkably outperforms state-of-the-art methods across four benchmark datasets and three label settings. It achieves a groundbreaking reduction in the Mean Absolute Error on the JHU-Crowd dataset with only 5\% of the data labeled, lowering it to 67.0, a decrease of 12.4\%. Furthermore, the method demonstrates exceptional generalization capabilities in cross-dataset Domain Generalization settings, even surpassing the performance of fully-supervised approaches, setting a new standard for excellence.
\end{itemize}

\begin{figure*}[ht]
    \centering
    \vspace{-0.3 cm}
    \includegraphics[width=0.95\linewidth]{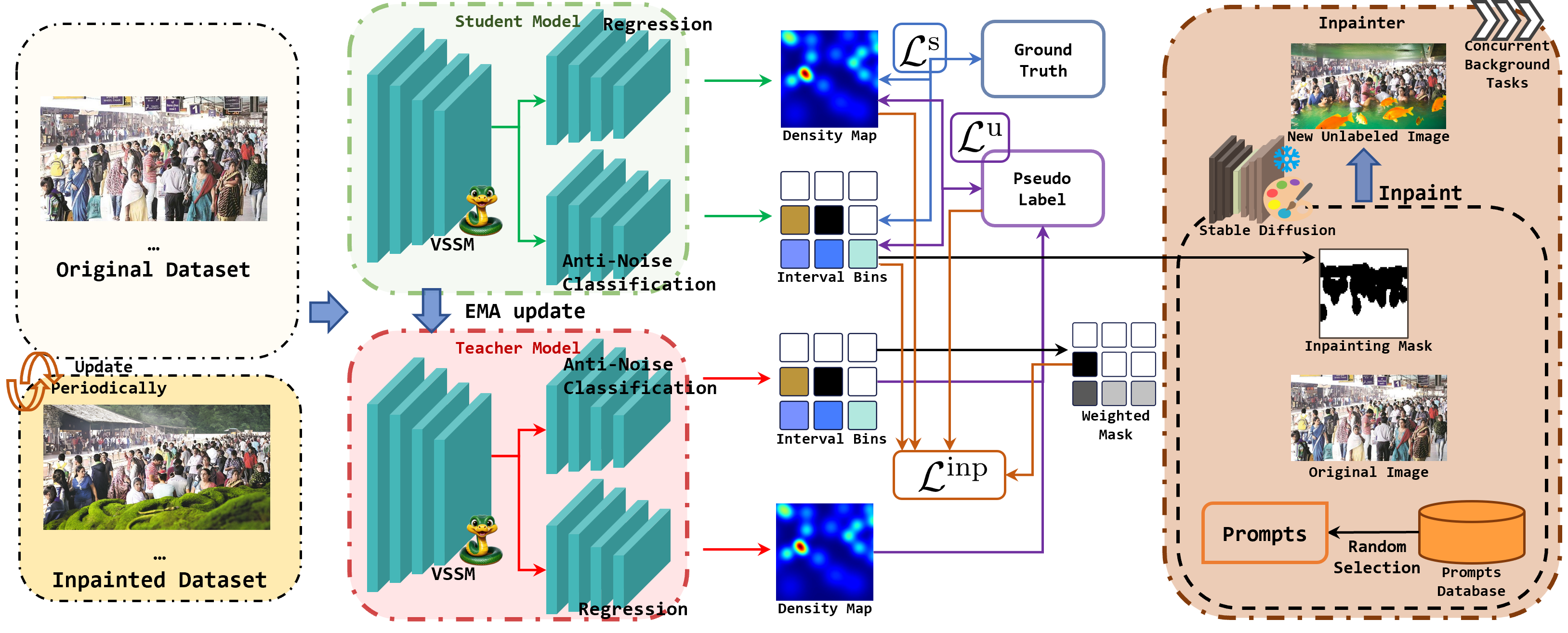}
    \caption{The overall framework of our method TMTB. TMTB contains the Mean Teacher framework for semi-supervised learning, VSSM as the backbone, a classification branch for predicting masks, and an inpainter for inpainting augmentation. The inpainting process is conducted periodically, which generates or updates the inpainted images. For filtering out unreliable regions, the teacher model generates the weighted mask based on the inconsistency level of the inpainted image, which is applied in $\mathcal{L}^{\text{inp}}$. The teacher model generates pseudo-labels for unlabeled data (inpainted images included), and the student model predicts on labeled data and strongly augmented unlabeled data (inpainted images included).}
    \label{fig:method_overview}
    \vspace{-0.2 cm}
\end{figure*}

\section{Releted Works}

\noindent\textbf{Semi-supervised Crowd Counting.} 
Fully-supervised crowd counting methods\cite{shanghaiAB,ucfqnrf,Ma_2019_ICCV,song2021rethinking,Han_2023_ICCV,chen2024effectivenesssimplifiedmodelstructure,Peng_2024_CVPR,Ranasinghe_2024_CVPR} maintain the best performance in crowd counting tasks. 
However, fully-supervised methods require a large quantity of labeled data, which is expensive to collect and annotate. 

Therefore, semi-supervised crowd counting methods have been proposed to address the scarcity of labeled data. 
The Mean Teacher~\cite{mean_teacher}, a successful semi-supervised learning framework that has been widely used in classification~\cite{fixmatch,NEURIPS2021_995693c1,DBLP:journals/tnn/DuanZQWZSG24}, segmentation~\cite{Wang_2022_CVPR,bai2023bidirectionalcopypastesemisupervisedmedical} and regression~\cite{NEURIPS2018_9d28de8f,Yin_2022_CVPR,Dai_Li_Cheng_2023}, is the most popular method in semi-supervised crowd counting. 
L2R~\cite{l2r} leveraged unlabeled data through a learning-to-rank framework. 
IRAST~\cite{irast_eccv} introduced surrogate tasks, self-learning inter-related binary segmentation tasks, to improve performance. Moreover, MTCP~\cite{mtcp_tnnls} applied multiple auxiliary tasks into representation learning, which also benefit performance. 
DACount~\cite{density_agency} employed a Transformer~\cite{vaswani2023attentionneed} to refine foreground features. 
CU~\cite{Calibrating} investigated a supervised uncertainty estimation strategy to better utilize unlabeled data. 
SALCrowd~\cite{zhang2024boosting} proposed a method to select unlabeled data based on intra-scale uncertainty and inter-scale inconsistency metrics. 
MRC-Count~\cite{mrc-crowd2024} incorporated a fine-grained density classification task, proving the effectiveness of the classification task in crowd counting. 
Most previous approaches \cite{l2r,irast_eccv,Calibrating,zhang2024boosting,mrc-crowd2024} are based on CNNs or partially \cite{density_agency} based on Transformers, which are weak in modeling long-range dependencies. 
When facing extremely dense crowd scenes, the performance of these methods is not satisfactory.

\noindent\textbf{Generative Data Augmentation.} 
Deep learning models tend to overfit when the quantity of data is limited, which harms the generalization ability of models. 
Therefore, data augmentation was proposed to address this issue. 
For image-based tasks, basic data augmentation methods include random cropping, flipping, rotation, scaling, color jittering, and noise adding. 
Some advanced data augmentation methods, such as CutMix~\cite{yun2019cutmixregularizationstrategytrain}, Mixup~\cite{zhang2018mixupempiricalriskminimization}, and AutoAugment~\cite{cubuk2019autoaugmentlearningaugmentationpolicies}, have been proposed. 
However, these methods are not suitable for crowd counting tasks. 
Both the density-based and point-based crowd counting methods are sensitive to the spatial structure of the crowd, and the above methods may destroy the spatial structure of the crowd. 
The damage to the integrity of the crowd will lead to a decrease in model performance. 
We study this in \Cref{tab:aug_compare}.

Generative Data Augmentation \cite{kingma2013auto,goodfellow2020generative,ho2020denoising,denoising_diffusion,ramesh2021zero} is the process of creating new synthetic data points that can be added to existing datasets. 
This unique approach is widely used to improve the performance of models by increasing the quantity and diversity of training data while maintaining the quality of the data. 
In \cite{zhang2024gettinglesslargelanguage}, queries-only data was used to improve the multilingual task, showing the effectiveness of generative data augmentation. 
FreeMask~\cite{freemask} resorted to synthetic images from generative models to ease the burden of both data collection and annotation procedures, and achieves comparable performance to real-data-based methods. 
$\text{G-DAUG}^\text{C}$~\cite{yang2020generative} generated synthetic examples using pretrained language models, and selects the most informative and diverse set of examples for data augmentation, to achieve more accurate and robust learning in low-resource setting. 
DiffusionMix~\cite{Islam_2024_CVPR} employed an inpaint-and-mix strategy to generate label-preserved data for classification task. However, it requires pre-defined binary masks to concatenate the original and generated images, which is not available in crowd counting since the position and shape of the crowd are unpredictable. 
Compared to existing works, our contribution is to fill the gap in effective augmentation methods within the field of pixel-level predition tasks.

\section{Method}

We propose a semi-supervised crowd counting method that enhances the dataset using a diffusion-based model's inpainting technique and utilizes a VSSM (Visual State Space Model) with Anti-Noise classification branch to improve data diversity and model performance. 

The overall framework of our method, TMTB, is shown in \Cref{fig:method_overview}. Our method is built on the Mean Teacher framework, which is widely employed in semi-supervised learning. 
The teacher model shares the same architecture as the student model, and its weights are updated using the student model's weights through the Exponential Moving Average (EMA) strategy. 
VMamba~\cite{liu2024vmambavisualstatespace}, a novel VSSM (Visual State Space Model), serves as the backbone, replacing the CNN and Transformer. 
The image features extracted from VMamba are then fed into two independent branches: the counting branch and the classification branch, for predictions. 
The counting branch predicts a density map of the image, while the classification branch predicts the indices of pre-defined count interval bins. 
We build a module, Inpainter, designed to periodically applying background inpainting augmentation to the training set.

We first describe our novel inpainting augmentation in \Cref{sec:inpainter}, then we discuss our network architecture in \Cref{sec:network}, and finally, we present our loss functions in \Cref{sec:losses}.

\subsection{Inpainting Augmentation}

\label{sec:inpainter}

Unlike classification or regression tasks, the ground truth density maps are generated by point-level annotations. 
The state-of-the-art data augmentation techniques, Mixup~\cite{zhang2018mixupempiricalriskminimization}, CutMix~\cite{yun2019cutmixregularizationstrategytrain}, BCP~\cite{bai2023bidirectionalcopypastesemisupervisedmedical}, etc., are not applicable to crowd counting, since the augmentation may cause damage to ground truth density maps. 
We introduce the inpainting technique of diffusion-based models \cite{denoising_diffusion,latent_diffusion} as a novel data augmentation method with the filtering capability of unreliable regions, which generates new samples, thereby expanding the diversity of dataset.

For the inpainting mask $M^{\text {inp }}$, the count interval bins predicted by the classification branch are used to separate the foreground and the background of image $x$. 
We generate the positive text prompt randomly to achieve more diverse unlabeled dataset, but the negative text prompt is kept the same for all samples. 
The inpainting process is conducted periodically during the training process using a preset time interval $T^{\text{inp}}$:
\begin{equation}
    \label{eq:inpaint_mask}
    \begin{split}
        M^{\text {inp }} & = \mathbf{0} \left( \arg\max \left( \hat{p} \right) \right), \\
        x^{\text {inp }} & = \mathcal{SD} \left( x, M^\text{inp}, r^\text{pos}, r^\text{neg} \right),
    \end{split}
    \vspace{-0.2 cm}
\end{equation}
where $\hat{p}_{i}$ is the predicted probability distribution from the classification branch, $\mathbf{0} \left( \cdot \right)$ is the indicator function, $x^{\text {inp }}$ is the new image with the background inpainted, $\mathcal{SD \left( \cdot ,\cdot,\cdot,\cdot \right)}$ represents Stable Diffusion~\cite{latent_diffusion}, which is a mature, excellent, and cost-effective diffusion model, and $r^\text{pos}$ and $r^\text{neg}$ are the positive and negative text prompts, respectively.

\begin{figure}
    \centering
    \includegraphics[width=0.9\linewidth]{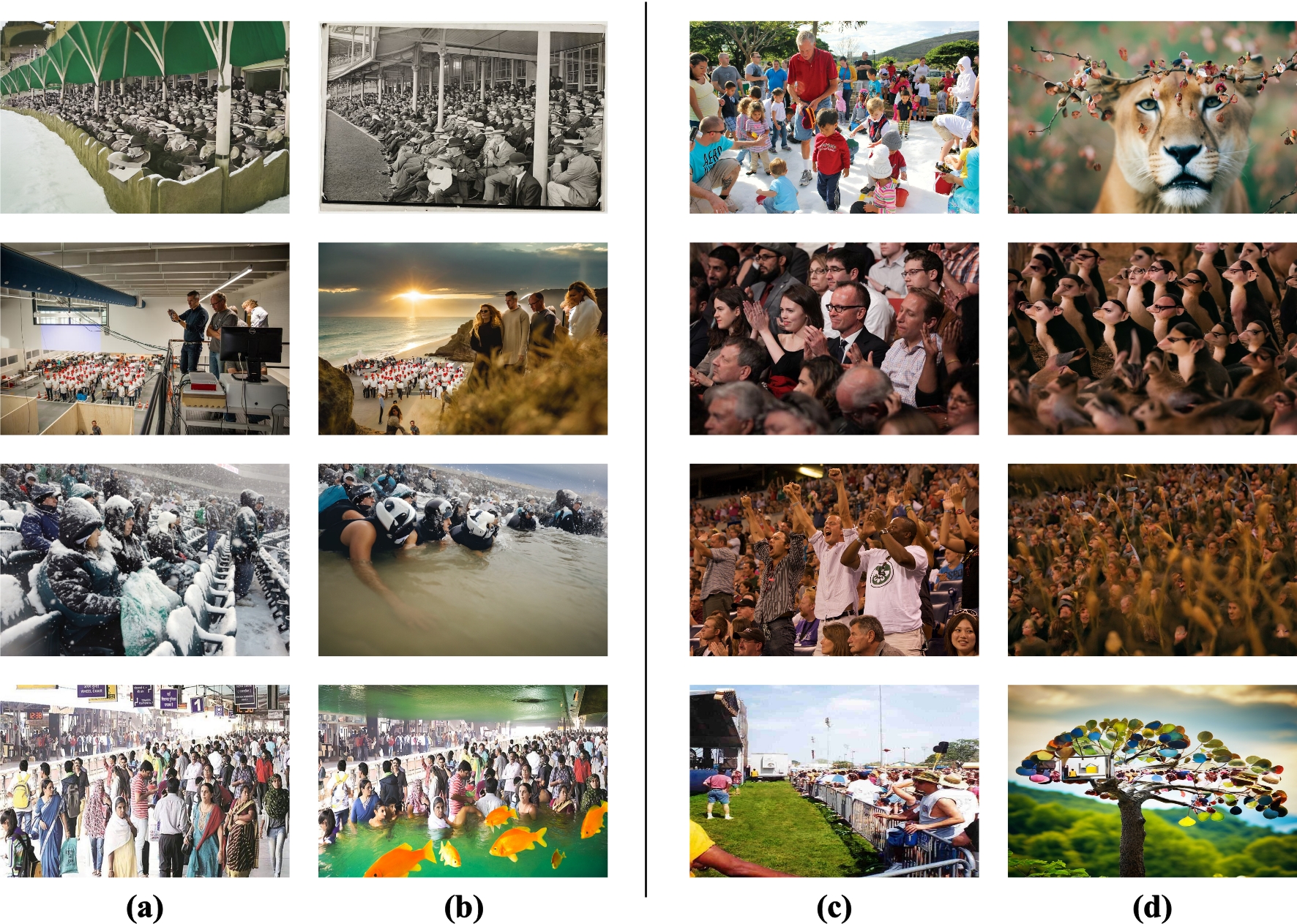}
    \vspace{-0.2 cm}
    \caption{Inpainting samples. The left part shows well-inpainted couples, while the right part shows poorly-inpainted couples. Column (a) and (c) are the original images, while column (b) and (d) are the inpainted images.}
    \label{fig:good_bad_samples}
    \vspace{-0.4 cm}
\end{figure}

\Cref{fig:good_bad_samples} shows examples of both good and poor inpainting samples. 
All new samples are considered unlabeled, as the inpainting process may add additional humans for better coherence. 
For well-inpainted samples, the foreground regions are well preserved, while the background regions are inpainted with different patterns, thus introducing diversity to the dataset.
Even though the inpainting process is not perfect, the model can still benefit from the augmented dataset.
However, failures may sometimes occur, potentially leading to the model learning from noisy inputs. 
The heads of humans and other features in the unreliable regions of poorly-inpainted images are fragmented and indistinct, rendering them difficult to recognize. 
It is necessary to \textbf{filter out} these \textbf{unreliable regions}. 
The EMA model calculates the inconsistency level of the inpainted image by the difference in the classification branch between the weakly augmented version and the strongly augmented version. 
The inconsistency level and the weighted mask are defined as follows:
\begin{equation}
    \label{eq:inpaint_weight}
    \begin{split}
        M_l^{\text{incon}} & = \mathbf{0} \left( \left| \left| \hat{p^s} - \hat{p^w} \right| - l \right| \right), \\
        \omega_l^t & = \operatorname{softmax} \left(e^{-l \cdot t / T^{\text{inpw}}}\right), \\
        M^{\text{incon}} & = \sum_{l=0}^{L} M_l^{\text{incon}} \cdot \omega_l^t, \\
    \end{split}
    \vspace{-0.2 cm}
\end{equation}
where $\hat{p^s}$ and $\hat{p^w}$ are the predicted probability distributions of the classification branch of EMA model in the strong and weak augmentation version, respectively. $M_l^{\text{incon}}$ is the mask of inconsistency level $l$, $M^{\text{incon}}$ is the final weighted mask, $\omega_l^t$ is the weight of inconsistency level $l$ at time $t$, $T^{\text{inpw}}$ is the parameter controlling the weight decay speed, and $L$ is the number of inconsistency levels. The $L$ depends on the predefined count interval bins, and since we followed the interval bins settings in \cite{Wang_2021_ICCV}, the $L$ is set to $2$ for the balance of diversity and reliability. 
The weighted mask is then used to adjust the loss associated with the inpainted image.

\begin{figure*}
    \centering
    \vspace{-0.3 cm}
    \includegraphics[width=0.85\linewidth]{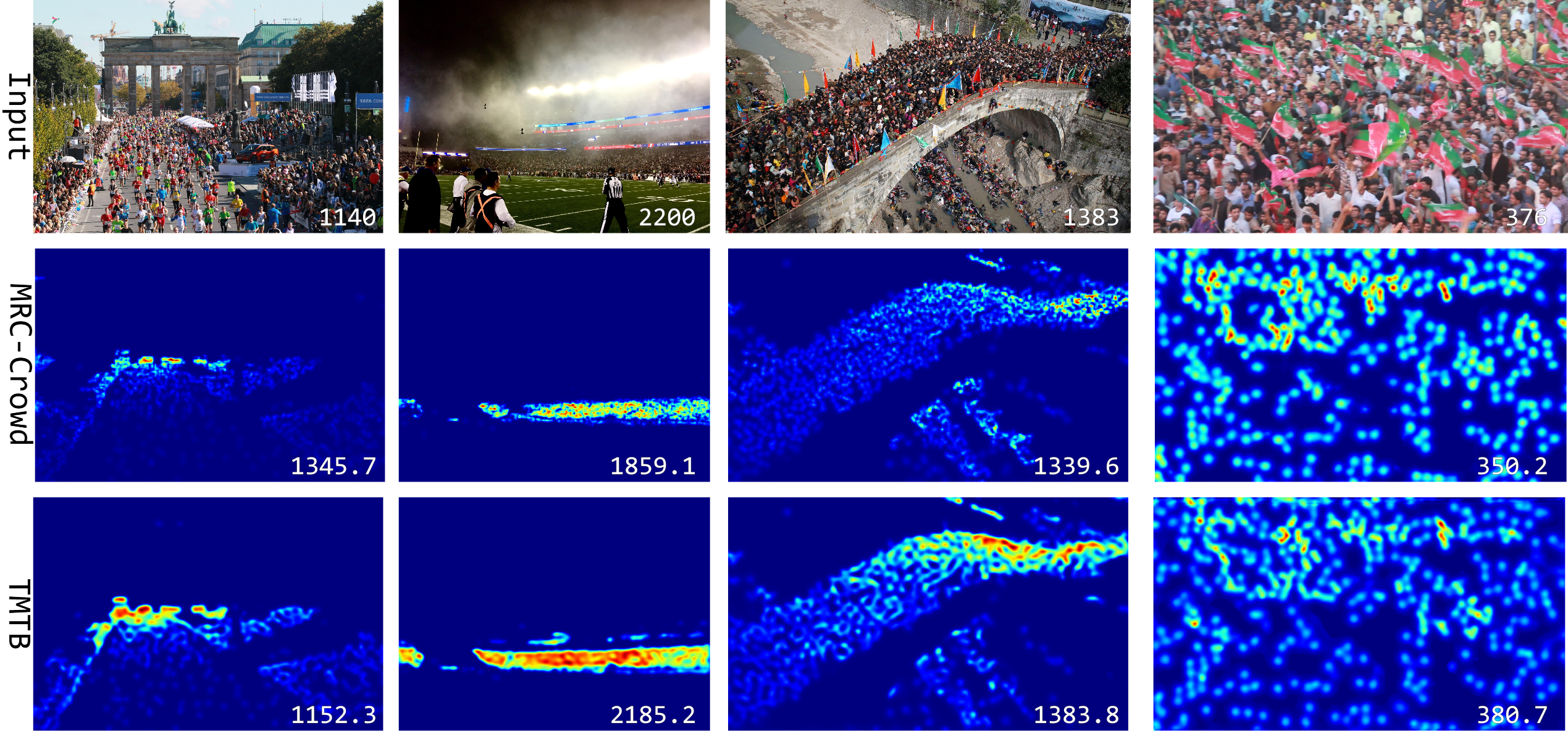}
    \caption{Visualizations of the predictions on the test set of the JHU-Crowd++ dataset. The first row shows the input images. The second row displays the predictions of the SOTA method MRC-Crowd~\cite{mrc-crowd2024}. The third row presents the predictions of our method TMTB. All models are trained with 5\% labeled data.}
    \label{fig:overall_results}
    \vspace{-0.4 cm}
\end{figure*}

\subsection{VSSM with Classification for Anti-Noise}

\label{sec:network}

\noindent\textbf{Visual State Space Models (VSSMs).} 
Existing crowd counting methods often encounter challenges such as congestion, blur, and occlusion in intricate scenes. 
CNN-based approaches tend to overfit to local details, while Transformer-based methods excel in global modeling but face heavy computational burdens due to their quadratic complexity in terms of image sizes. 
These limitations motivate us to introduce the VSSM, which captures comprehensive long-range information with linear time complexity, for the crowd counting task.

State Space Models (SSMs) map the input stimulation $ u(t) $ to the output response $ y(t) $ by means of a hidden state $ h(t)$. These processes are formulated with the following form:
\begin{equation}
    \label{eq:ssm}
    \begin{split}
        {h'}(t) & = {A} {h}(t) + {B} u(t), \\
        y(t)           & = {C} {h}(t) + {D} u(t),
    \end{split}
    \vspace{-0.4 cm}
\end{equation}
where ${A}$, ${B}$, ${C}$, and ${D}$ are the weighting parameters.

Linear time-invariant SSMs (\Cref{eq:ssm}) lack the ability to capture the contextual information, \cite{gu2024mambalineartimesequencemodeling} proposed a novel parameterization method for SSMs that integrates an input-dependent selection mechanism (referred to as S6).
VMamba~\cite{liu2024vmambavisualstatespace} replaces the S6 module, with the newly proposed 2D-Selective-Scan (SS2D) module which concurrently achieving global receptive fields, dynamic weights, and linear complexity. 
By adopting complementary 1D traversal paths, SS2D enables each pixel in the image to effectively integrate information from all other pixels in different directions, thereby facilitating the establishment of global receptive fields in the 2D space and generating more accurate density map. 
Details about the SS2D module could be found in \cite{liu2024vmambavisualstatespace}.
The features extracted by the final SS2D block are then fed into two branches: one for density map regression and the other for count interval classification, which is designed for less inexact and more accurate supervision.

\noindent\textbf{Exact: Regression Head.} 
The regression head is designed to predict the density map for the input image. The density map is then summed to obtain the exact final count value.
The regression head consists of up-sampling layers, convolutional layers, batch normalization layers, and ReLU activation layers. %After two up-sampling layers, the density map is up-sampled to $H/8\times W/8$. 
We followed MRC-Crowd~\cite{mrc-crowd2024} employing the loss function proposed in CUT~\cite{Qian_2022_BMVC}, which significantly boosts the performance in the foreground area. It could be expressed as:
\begin{equation} 
    \label{eq:loss_reg}
    \begin{split}
        \mathcal{L}_{\text {reg }} & =\frac{1}{J} \sum_{j=1}^{J}\left(1-\operatorname{SSIM}\left(D_{j}\left(\hat{y} \odot M^{g t}\right),\right.\right. \\ 
        & \left.\left. D_{j}\left(y^{g t} \odot M^{g t}\right)\right)\right) +\alpha \cdot \mathcal{L}_{T V}\left(\hat{y}, y^{g t}\right),
    \end{split}
\end{equation}
where $\operatorname{SSIM \left( \cdot,\cdot \right)}$ is the structural similarity index measure, and $D_j$ denotes the average pooling operation that downsamples an image to $1/2^j$ of its original size. 
The operation $\odot$ denotes the Hadamard product, and $M^{gt}$ is the binary mask that distinguishes the dense and sparse regions based on a predefined threshold $\tau$. 
It can be obtained using the indicator function $\mathbf{1}$ as $M^{gt} = \mathbf{1}\left(y^{gt} > \tau\right)$. 
$\alpha$ is a hyper-parameter that balances the two terms. $J$ and $\alpha$ are set to $3$ and $0.01$, respectively. 
$\mathcal{L}_{TV}$ is the total variation loss.
 
\noindent\textbf{Accurate: Anti-Noise Classification Head.} 
Many previous works~\cite{olmschenk2019improving,Wan_2019_ICCV,Ma_2019_ICCV,Bai_2020_CVPR} have identified the problem of inaccurate ground truth annotations in crowd counting datasets, which render the ground truth density maps noisy and inaccurate.
In this way, the model may learn from the noisy ground truth and generate inaccurate density maps. 
Inspired by previous works~\cite{mrc-crowd2024,Wang_2021_ICCV,density_agency}, we employ the Anti-Noise classification task as an auxiliary task, which is more robust in predicting inexactly and accurately. 
The Anti-Noise classification branch is supervised by the inexact but accurate count interval bins rather than the exact but maybe inaccurate density maps.
Furthermore, we use the prediction of the this branch to generate reliable masks for inpainting samples, and we discuss this in \Cref{sec:inpainter}.
The goal of the Anti-Noise classification branch is to predict the accurate count interval bins, which are predefined by \cite{Wang_2021_ICCV}.

The Anti-Noise classification head consists of an up-sampling layer, a convolutional layer, a ReLU activation layer and a final convolutional layer.
The standard cross-entropy loss is applied. In the case of labeled data, the loss is defined as:
\begin{equation}
    \label{eq:loss_cls}
    \mathcal{L}_{\text {cls }}=\frac{1}{N} \sum_{i}^{N} \mathcal{H}\left(p_{i}^{g t}, \hat{p}_{i}\right),
\end{equation}
where $\mathcal{H}$ is the cross-entropy loss function, $p_{i}^{gt}$ is the ground truth probability distribution, and $\hat{p}_{i}$ is the predicted probability distribution.

\subsection{Losses}

\label{sec:losses}

In order to make full use of the labeled data $L = \left\{ \left( x_i^l, {DM}_i^{gt}  \right) \right\}_{i=1}^{N_l}$ and unlabeled data $ U = \left\{ \left( x_i^u \right)   \right\}_{i=1}^{N_u} $, we employ a common semi-supervised learning strategy widely used in crowd counting. 
The labeled samples are used to directly train the student model, while the unlabeled samples are used for consistency regularization. 
For labeled data, the supervised loss $\mathcal{L}^{\text{s}} = \mathcal{L}_{\text{reg}}^{\text{s}} + \mathcal{L}_{\text{cls}}^{\text{s}}$ is a combination of the regression loss (\Cref{eq:loss_reg}) and the classification loss (\Cref{eq:loss_cls}). 
Given the absence of a confidence score for the predicted regression map, we choose to apply consistency regularization to all unlabeled samples, rather than developing an algorithm to assess prediction uncertainty. 
The EMA teacher model generates pseudo-labels based on the prediction of unlabeled data with weak augmentation. 
The student model makes predictions based on the strong augmentation version. 
The error loss between the two predictions is calculated as the consistency loss. 

Despite the common strong augmentation, we also introduce a patch-aligned random masking strategy \cite{xie2022simmim} as a kind of strong augmentation. 
The patch-aligned random masking strategy randomly masks out some patches of various sizes in the image. The student model is required to predict based on masked images, while the teacher model is able to access the full image. 
The semi-supervised loss is defined as:
\begin{equation}
    \label{eq:loss_unsup}
    \mathcal{L}^{\text{u}} = \operatorname{MAE} \left( \hat{y}^{st}, \hat{y}^{ema} \right) + \operatorname{MAE} \left( \hat{p}^{st}, \hat{p}^{ema} \right),
    \vspace{-0.1 cm}
\end{equation}
where $\hat{y}^{st}$ and $\hat{y}^{ema}$ are the predicted density maps of the student model and the teacher model, respectively. $\hat{p}^{st}$ and $\hat{p}^{ema}$ are the predicted count interval bins of the student model and the teacher model, respectively. 

Consistency loss is also applied to the inpainted images. 
However, to avoid the model learning from noisy regions with low quality, the loss is adjusted by the weighted mask of inconsistency level. 
The inpainting loss is defined as:
\begin{equation}
    \label{eq:loss_inpaint}
    \mathcal{L}^{\text{inp}} = M^{\text{incon}} \times \text{MAE} \left( \hat{y}^{st}, \hat{y}^{ema} \right) + M^{\text{incon}} \times \text{MAE} \left( \hat{p}^{st}, \hat{p}^{ema} \right),
\end{equation}
where $M^{\text{incon}}$ is the weighted mask of inconsistance level mentioned in \Cref{eq:inpaint_weight}.

The final loss is a combination of the supervised, semi-supervised, and inpainting losses:
\begin{equation}
    \label{eq:loss_final}
    \mathcal{L} = \mathcal{L}^{\text{s}} + \lambda_{w} \cdot \mathcal{L}^{\text{u}} + \lambda_{w} \cdot \mathcal{L}^{\text{inp}},
    \vspace{-0.2 cm}
\end{equation}
where $\lambda_{w}$ is the warm-up weight for the semi-supervised loss and the inpainting loss.

\section{Experiments}

\begin{table*}
    \caption{Comparisons with the SOTA methods on main public datasets. The best results are highlighted in bold, and the second-best results are underlined.}
    \label{tab:overall_results}
    \begin{center}
    \vspace{\tablegap}
    \resizebox{0.9\textwidth}{!}{
      \begin{tabular}{l|l|c|cc|cc|cc|cc}
\toprule
\multirow{2}{*}{Methods} & \multirow{2}{*}{Venue}& Labeled & \multicolumn{2}{c|}{JHU-Crowd++} & \multicolumn{2}{c|}{UCF-QNRF} & \multicolumn{2}{c|}{ShanghaiTech A} & \multicolumn{2}{c}{ShanghaiTech B} \\
\cline{4-11}
 & & Percentage  & MAE $\downarrow$  & RMSE $\downarrow$   &  MAE $\downarrow$  & RMSE $\downarrow$   & MAE $\downarrow$  & RMSE $\downarrow$  & MAE $\downarrow$  & RMSE $\downarrow$  \\
\midrule
MT~\cite{mean_teacher}          & NIPS'17   & 5\%  & 101.5 & 363.5 & 172.4 & 284.9 & 104.7 & 156.9 & 19.3  & 33.2  \\
L2R~\cite{l2r}         & CVPR'18   & 5\%  & 101.4 & 338.8 & 160.1 & 272.3 & 103.0 & 155.4 & 20.3  & 27.6  \\
DACount~\cite{density_agency}     & ACM MM'22 & 5\%  & 82.2  & 294.9 & 120.2 & 209.3 & 85.4  & 134.5 & 12.6  & 22.8  \\
OT-M~\cite{lin2023optimal}        & CVPR'23   & 5\%  & 82.7  & 304.5 & 118.4 & 195.4 & 83.7  & 133.3 & 12.6  & 21.5  \\
MRC-Crowd~\cite{mrc-crowd2024}   & T-CSVT'24 & 5\%  & \underline{76.5}  & \underline{282.7} & \underline{101.4} & \underline{171.3} & 74.8  & \textbf{117.3} & \underline{11.7}  & \textbf{17.8}  \\
\textbf{TMTB}      &  \textbf{Ours}   & 5\%  & \textbf{67.0}  & \textbf{252.2} & \textbf{96.3}  & \textbf{163.4} & \textbf{72.4}  & \underline{126.4} & \textbf{10.6}  & \underline{19.6}  \\
\midrule
MT~\cite{mean_teacher}          & NIPS'17   & 10\% & 90.2  & 319.3 & 145.5 & 250.3 & 94.5  & 156.1 & 15.6  & 24.5  \\
L2R~\cite{l2r}         & CVPR'18   & 10\% & 87.5  & 315.3 & 148.9 & 249.8 & 90.3  & 153.5 & 15.6  & 24.4  \\
DACount~\cite{density_agency}     & ACM MM'22 & 10\% & 75.9  & 282.3 & 109.0 & 187.2 & 74.9  & 115.5 & 11.1  & 19.1  \\
MTCP~\cite{mtcp_tnnls}       & T-NNLS'23 & 10\% & 88.1  & 318.7 & 124.7 & 206.3 & 81.3  & 130.5 & 14.5  & 22.3  \\
OT-M~\cite{lin2023optimal}        & CVPR'23   & 10\% & 73.0  & 280.6 & 113.1 & 186.7 & 80.1  & 118.5 & 10.8  & 18.2  \\ 
CU~\cite{Calibrating}    & ICCV'23   & 10\% & 74.9  & 281.7 & 104.0 & 164.3 & 70.8  & 116.6 & \textbf{9.7}  & 17.7  \\
SALCrowd~\cite{zhang2024boosting} & ACM MM'24 & 10\% & \underline{69.7} & \underline{263.5} & 106.7 & 171.3 & 69.7 & 114.5 & \textbf{9.7} & \underline{17.5} \\
MRC-Crowd~\cite{mrc-crowd2024}   & T-CSVT'24 & 10\% & 70.7  & \textbf{261.3} & \underline{93.4}  & \textbf{153.2} & \underline{67.3}  & \textbf{106.8} & 10.3  & 18.2  \\
\textbf{TMTB}      &  \textbf{Ours}   & 10\% & \textbf{66.3}  & 265.1 & \textbf{91.7}  & \underline{156.3} & \textbf{65.7}  & \underline{110.4} & \underline{9.8}  & \textbf{17.2}  \\
\midrule
MT~\cite{mean_teacher}          & NIPS'17   & 40\% & 121.5 & 388.9 & 147.2 & 247.6 & 88.2  & 151.1 & 15.9  & 25.7  \\
L2R~\cite{l2r}         & CVPR'18   & 40\% & 123.6 & 376.1 & 145.1 & 256.1 & 86.5  & 148.2 & 16.8  & 25.1  \\
SUA~\cite{sua_iccv}        & ICCV'21   & 40\% & 80.7  & 290.8 & 130.3 & 226.3 & 68.5  & 121.9 & 14.1  & 20.6  \\
DACount~\cite{density_agency}     & ACM MM'22 & 40\% & 65.1  & 260.0 & 91.1  & 153.4 & 67.5  & 110.7 & 9.6   & 14.6  \\
OT-M~\cite{lin2023optimal}        & CVPR'23   & 40\% & 72.1  & 272.0 & 100.6 & 167.6 & 70.7  & 114.5 & 8.1   & 13.1  \\
SALCrowd~\cite{zhang2024boosting} & ACM MM'24 & 40\% & \underline{62.0} & 255.1 & 95.0 & 167.0 & \textbf{60.7} & \underline{97.2} & 7.9 & \textbf{12.7} \\
MRC-Crowd~\cite{mrc-crowd2024}   & T-CSVT'24 & 40\% & \textbf{60.0}  & \textbf{227.3} & \textbf{81.1}  & \textbf{131.5} & 62.1  & \textbf{95.5}  & \underline{7.8}   & 13.3  \\
\textbf{TMTB}      &  \textbf{Ours}   & 40\% & \textbf{60.0}  & \underline{240.3} & \underline{81.4}  & \underline{140.6} & \underline{60.8}  & 99.0  & \textbf{7.5}  & \underline{12.9}  \\
\bottomrule
     \end{tabular}}
     \vspace{\tablegap}
    \end{center}
\end{table*}

\label{sec:exp}

We evaluate our method on four public datasets:
\begin{itemize}
    \item \textbf{JHU-Crowd++}~\cite{sindagi2020jhu-crowd++} is a comprehensive dataset with 4,372 images and 1.51 million annotations, collected under a variety of scenarios and environmental conditions.
    \item \textbf{UCF-QNRF}~\cite{ucfqnrf} contains 1535 images. The dataset has a greater number of high-count crowd images and annotations, and a wider variety of scenes containing the most diverse set of viewpoints, densities and lighting variations.
    \item \textbf{ShanghaiTech A \& B}~\cite{shanghaiAB} are two datasets containing nearly 1,200 images and 330,000 labeled heads. 
\end{itemize}

To achieve fair comparisons, we follow the same semi-supervised experimental protocol as previous works~\cite{density_agency,lin2023optimal,mrc-crowd2024}, including the 5\%, 10\%, and 40\% labeled data settings and the split of labeled and unlabeled data. 
The evaluation metrics are Mean Absolute Error (MAE) and Root Mean Squared Error (RMSE). The detailed definitions of them are provided in the supplementary material.

For some excessively high-resolution images, we scale them down, ensuring the longest side does not exceed 1,920 pixels. 
For the ground truth density map, we use geometry-adapted Gaussian kernels described in \cite{li2018csrnetdilatedconvolutionalneural} for all datasets except the ShanghaiTech B dataset, and we fix the size of the kernel to $4$. 
When training, the input images are randomly cropped to $512 \times 512$ pixels for all datasets except for ShanghaiTech A. The inputs from ShanghaiTech A are cropped to $256 \times 256$ pixels instead. 
The count interval bins, which should be predicted by the classification branch, are set to the same as those in \cite{Wang_2021_ICCV}. 
The AdamW optimizer~\cite{adam} with a learning rate of $10^{-5}$ is used to optimize the model. The weight decay is set to $ 10^{-4}$. 
The downsample factor is set to $8$. The batch size is set to $8$. In 5\% and 10\% labeled data settings, the ratio of labeled data to unlabeled data is set to $2:6$. In the 40\% labeled data setting, the ratio is set to $4:4$. 
As we mentioned in \Cref{sec:losses}, we use the patch-aligned random masking strategy \cite{xie2022simmim} as a form of strong augmentation. The masked patch size and the masked region ratio are set to $32 \times 32$ and $0.3$, respectively. The ablation details can be found in \cite{mrc-crowd2024}. 
The teacher model is updated by the exponential moving average with a decay rate of $0.97$. The weights of all losses are set to $1.0$. For $\mathcal{L}^{\text{u}}$ and $\mathcal{L}^{\text{inp}}$, we ramp the weights from $0$ to $1.0$ in the first 20 epochs. The max accepted inconsistency level $L$ is set to $2$, and the hyperparameter $T^{\text{inpw}}$ is set to $100$. The ablation study of $T^{\text{inpw}}$ is conducted in \Cref{sec:abl}. Each training process coordinates with two inpainting background processes. The inpainting period $T^{\text{inp}}$ is set to $80$ epochs.

\subsection{Main Results}

In our main experiments, the proposed TMTB demonstrates superior performance compared to state-of-the-art methods across four public datasets with three labeled data percentages. 
The results are shown in \Cref{tab:overall_results}. 

Notably, when utilizing only 5\% labeled data, TMTB achieves the lowest MAEs of 67.0 on JHU-Crowd++, 96.3 on UCF-QNRF, 72.4 on ShanghaiTech A, and 10.6 on ShanghaiTech B, demonstrating the best performance. 
The result on JHU-Crowd++ not only surpasses the performance of SOTA methods with the same amount of labeled data but also demonstrates that TMTB remains superior even at the 10\% label setting, highlighting its efficient label utilization and exceptional accuracy.
With a 10\% labeled data setting, TMTB continues to outstand, obtaining a leading MAE of 66.3 on JHU-Crowd++, 91.7 on UCF-QNRF, and 65.7 on ShanghaiTech A, further demonstrating its generalization ability and effectiveness. 
At 40\% labeled data, TMTB achieves competitive results, matching the lowest MAE of 60.0 on JHU-Crowd++, and decreasing the MAE to a lowest of 7.5 on ShanghaiTech B, maintaining strong performance across the other datasets.
These results underscore the capability of TMTB to achieve low error with minimal labeled data, offering substantial improvements over SOTA methods.

\begin{table}
    \caption{An study of common various augmentation strategies on ShanghaiTech-A dataset using 10\% labeled data.}
    \label{tab:aug_compare}
    \vspace{\tablegap}
    \begin{center}
        \resizebox{0.75\linewidth}{!}{
        \begin{tabular}{l|cc}
        \toprule
            Augmentation & MAE $\downarrow$ & RMSE $\downarrow$ \\
        \midrule
            Mixup~\cite{zhang2018mixupempiricalriskminimization} & 285.16 & 431.83 \\
            Mixup~\cite{zhang2018mixupempiricalriskminimization} (background only) & 72.00 & 116.85 \\
            CutMix~\cite{yun2019cutmixregularizationstrategytrain} & 312.60 & 462.33 \\
            BCP~\cite{bai2023bidirectionalcopypastesemisupervisedmedical} & 70.99 & 121.83 \\
        \midrule
            \textbf{TMTB (Ours)} & 65.74 & 110.42 \\
        \bottomrule
        \end{tabular}}
    \end{center}
    \vspace{\tablegap}
\end{table}
\noindent\textbf{Augmentation Strategies.} 
In order to investigate the effectiveness of the proposed inpainting augmentation strategy, we compare it with some off-the-shelf augmentation strategies, on the ShanghaiTech A dataset with 10\% labeled data, including Mixup~\cite{zhang2018mixupempiricalriskminimization}, CutMix~\cite{yun2019cutmixregularizationstrategytrain}, and BCP~\cite{bai2023bidirectionalcopypastesemisupervisedmedical}. The results are shown in \Cref{tab:aug_compare}.

Mixup and CutMix both degrade the performance when applied to the whole image. 
The main reason is that the integrity of the foreground and background regions is destroyed, making the augmented images non-informative and hard to learn. These results demonstrate the importance of the integrity of the foreground and background regions in the crowd counting task. 
We then apply Mixup to the background region only, and the performance improves. This approach is somewhat similar to the inpainting augmentation, which enhances the background region while keeping the foreground region intact. However, the background regions are sampled from the domain of the original dataset, which may not be sufficiently diverse.
BCP is a recent birectical augmentation strategy designed for medical image segmentation. We employ it in the crowd counting task and find it can also improve the performance. However, the performance is not competitive enough when compared to SOTA methods.

\begin{table}
    \caption{Comparisons with SOTA single domain generalization methods on Q (UCF-QNRF) and A (ShanghaiTech-A) datasets. The best results are highlighted in bold, and the second-best results are underlined.}
    \label{tab:dg_compare}
    \vspace{\tablegap}
    \begin{center}
        \resizebox{\linewidth}{!}{
        \begin{tabular}{l|l|c|cc|cc}
        \toprule
            \multirow{3}{*}{Methods} & \multirow{3}{*}{Venue} & Source & \multicolumn{2}{c|}{\multirow{2}{*}{A $\rightarrow$ Q}} & \multicolumn{2}{c}{\multirow{2}{*}{Q $\rightarrow$ A}} \\
             & & Labeled & \multicolumn{2}{c|}{} & \multicolumn{2}{c}{} \\
            \cline{4-7}
             & & Percentage & MAE $\downarrow$ & RMSE $\downarrow$ & MAE $\downarrow$ & RMSE $\downarrow$ \\
        \midrule
        IBN~\cite{Pan_2018_ECCV} & ECCV'18 & 100\% & 280.2 & 561.0 & 105.9 & 174.6 \\
        SW~\cite{Pan_2019_ICCV} & ICCV'19 & 100\% & 285.5 & 431.0 & 102.4 & 168.8 \\
        ISW~\cite{Choi_2021_CVPR} & CVPR'21 & 100\% & 215.6 & 399.7 & 83.4 & 136.0 \\
        DG-MAN~\cite{Mansilla_2021_ICCV} & ICCV'21 & 100\% & 129..1 & 238.2 & - & - \\
        DCCUS~\cite{Du_Deng_Shi_2023} & AAAI'23 & 100\% & 119.4 & 216.6 & 67.4 & 112.8 \\
        MPCount~\cite{Peng_2024_CVPR} & CVPR'24 & 100\% & \underline{115.7} & \underline{199.8} & \underline{65.5} & \underline{110.1} \\
        \textbf{TMTB} & \textbf{Ours} & 40\% & \textbf{112.5} & \textbf{190.6} & \textbf{62.4} & \textbf{103.5} \\
        \bottomrule
        \end{tabular}}
    \end{center}
    \vspace{\tablegap}
\end{table}

\begin{table}
    \caption{Comparisons with SOTA domain adaptation methods on Q (UCF-QNRF) and A (ShanghaiTech-A) datasets. The best results are highlighted in bold, and the second-best results are underlined.}
    \label{tab:da_compare}
    \vspace{\tablegap}
    \begin{center}
        \resizebox{\linewidth}{!}{
        \begin{tabular}{l|l|cc|cc|cc}
        \toprule
            \multirow{3}{*}{Methods} & \multirow{3}{*}{Venue} & Source & Target & \multicolumn{2}{c|}{\multirow{2}{*}{A $\rightarrow$ Q}} & \multicolumn{2}{c}{\multirow{2}{*}{Q $\rightarrow$ A}} \\
             & & Labeled & Domain & \multicolumn{2}{c|}{} & \multicolumn{2}{c}{} \\
            \cline{5-8}
             & & Percentage & Visible & MAE $\downarrow$ & RMSE $\downarrow$ & MAE $\downarrow$ & RMSE $\downarrow$ \\
        \midrule
        RBT~\cite{RBT_mm20} & ACM MM'20 & 100\% & \Checkmark & 175.0 & 294.8 & - & - \\
        $\text{C}^2\text{MoT}$~\cite{c2mot} & ACM MM'21 & 100\% & \Checkmark & 125.7 & 218.3 & - & - \\
        FGFD~\cite{FGFD_mm22} & ACM MM'22 & 100\% & \Checkmark & 124.1 & 242.0 & 70.2 & 118.4 \\
        DAOT~\cite{daot_mm23} & ACM MM'23 & 100\% & \Checkmark & 113.9 & 215.6 & 67.0 & 128.4 \\
        FSIM~\cite{fsim_tmm} & T-MM'23 & 100\% & \Checkmark & \textbf{105.3} & \underline{191.1} & \underline{66.8} & \underline{111.5} \\
        \textbf{TMTB} & \textbf{Ours} & 40\% & \XSolidBrush & \underline{112.5} & \textbf{190.6} & \textbf{62.4} & \textbf{103.5} \\
        \bottomrule
        \end{tabular}}
    \end{center}
    \vspace{\tablegap}
\end{table}
\noindent\textbf{Domain Generalization \& Domain Adaptation.} 
We compare our method with state-of-the-art single domain generalization methods and obtain remarkable results in \Cref{tab:dg_compare}. 
The results show that our method outperforms the SOTA methods on both UCF-QNRF $\rightarrow$ ShanghaiTech-A and ShanghaiTech-A $\rightarrow$ UCF-QNRF tasks, utilizing only 40\% labeled data on the source domain instead of 100\%. 
Note that IBN, SW, and ISW are originally designed for image classification or segmentation, and are adapted to crowd counting and reported by MPCount~\cite{Peng_2024_CVPR}. 
The results demonstrate the effectiveness of VSSM and inpainting augmentation in our method.

The study on domain adaptation is also conducted, and the results are shown in \Cref{tab:da_compare}. 
Domain adaptation setting is easier than domain generalization, as the target domain is visible. 
It is surprising that our method achieves competitive results compared to SOTA methods, even though we only have 40\% labeled data available in the source domain and no access to the target domain. 
The comparisons again provide evidence of the generalization ability of TMTB.

\subsection{Ablation Study}
\label{sec:abl}

\begin{table}
    \caption{An ablation study on the effectiveness of various components on UCF-QNRF dataset using 5\% labeled data.}
    \label{tab:abl_components}
    \vspace{\tablegap}
    \begin{center}
        \resizebox{0.7\linewidth}{!}{
        \begin{tabular}{cccc|cc}
        \toprule
            VSSM & $\mathcal{L}^{\text{s}}$ & $\mathcal{L}^{\text{u}}$ & $\mathcal{L}^{\text{inp}}$ & MAE $\downarrow$ & RMSE $\downarrow$ \\
        \midrule
            \XSolidBrush & $\mathcal{L}_{\text{reg}}^\text{s}$ &  &  & 115.35 & 188.19 \\
            \Checkmark & $\mathcal{L}_{\text{reg}}^\text{s}$ &  &  & 110.41 & 185.52 \\
            \Checkmark & $\mathcal{L}_{\text{reg}}^\text{s} + \mathcal{L}^\text{s}_{\text{cls}}$ &  &  & 105.88 & 183.87 \\
            \Checkmark & $\mathcal{L}_{\text{reg}}^\text{s} + \mathcal{L}^\text{s}_{\text{cls}}$ & \Checkmark &  & 103.58 & 178.97 \\
            \Checkmark & $\mathcal{L}_{\text{reg}}^\text{s} + \mathcal{L}^\text{s}_{\text{cls}}$ & \Checkmark & $\omega_l$ & 97.21 & 166.06 \\
            \Checkmark & $\mathcal{L}_{\text{reg}}^\text{s} + \mathcal{L}^\text{s}_{\text{cls}}$ & \Checkmark & $\omega_l^t$ & 96.33 & 163.42 \\
        \bottomrule
        \end{tabular}}
    \end{center}
    \vspace{\tablegap}
\end{table}
\noindent\textbf{Components.} 
We conduct an ablation study to analyze the effectiveness of various components in our method on the UCF-QNRF dataset with 5\% labeled data. The results are shown in \Cref{tab:abl_components}. 
The VSSM is the core component. Compared to OT-M~\cite{lin2023optimal}, the VSSM with supervised loss $\mathcal{L}^{\text{s}}$ improves the performance by decreasing the MAE by 10.57\%. 
After adding the semi-supervised loss $\mathcal{L}^{\text{u}}$, the performance is further improved, showing the effectiveness of our semi-supervised learning strategy. 
Next, we introduce our novel inpainting augmentation strategy and its mask of inconsistency level $M^\text{uncon}_l$ but with fixed weights $ \omega_l$. The performance is further improved, and the MAE of $97.21$ surpasses the SOTA method MRC-Crowd~\cite{mrc-crowd2024} at $101.4$.
Finally, we employ time-based weights $\omega_l^t$ for the inpainting loss, and the performance is significantly improved, achieving the lowest MAE of $96.33$ and RMSE of $163.42$.

\noindent\textbf{Weight Decay Speed.} 
We study the impact of the weight decay speed $T^{\text{inpw}}$ on the JHU-Crowd++ dataset with 5\% labeled data. The detailed influence of $T^{\text{inpw}}$ on the MAE and RMSE is shown in supplementary material. 
We train and test TMTB with different $T^{\text{inpw}}$ values ranging from $80$ to $120$. 
The results show that both MAE and RMSE decrease as $T^{\text{inpw}}$ increases from $80$ to $100$. 
However, when $T^{\text{inpw}}$ is greater than $100$, the performance begins to degrade. 
Therefore, we set $T^{\text{inpw}}$ to $100$ in experiments.%, which achieves the best performance.

\section{Conclusion}

In the literature of semi-supervised crowd counting, a number of methods based on pseudo-labeling have been explored. Despite their success, these methods still struggle to achieve satisfactory performance in extremely crowded, low-light, and adverse weather scenarios. In this work, we reveal that the lack of suitable data augmentation methods and the weakness of capturing global context information are bottlenecks. 
Therefore, we design a novel inpainting augmentation method, Inpainting Augmentation that works better for crowd counting, enabling tasting more. 
Moreover, we introduce the Visual State Space Model to capture long-range dependencies in crowd counting, enabling tasting better. 
Our proposed framework, TMTB, achieves superior performance compared to existing methods.

{
    \clearpage
    \small
    \bibliographystyle{ieeenat_fullname}
    \bibliography{main}
}


\end{document}